\documentclass[english]{article}

\usepackage[T1]{fontenc}
\usepackage[latin9]{inputenc}
\usepackage{babel}
\usepackage{amsmath}
\usepackage{amssymb}
\usepackage{graphicx}
\usepackage[unicode=true,pdfusetitle,
 bookmarks=true,bookmarksnumbered=false,bookmarksopen=false,
 breaklinks=false,pdfborder={0 0 1},backref=section,colorlinks=false]
 {hyperref}
\usepackage{breakurl}

\makeatletter

\providecommand{\tabularnewline}{\\}

\usepackage{nips13submit_e}\usepackage{times}\usepackage{amsfonts}\usepackage{url}\title{Low-Rank Approximations for Conditional Feedforward Computation in Deep Neural Networks}

\author{
Andrew S. Davis \\
Department of Electrical Engineering and Computer Science \\
University of Tennessee \\
\texttt{andrew.davis@utk.edu} \\
\And
Itamar Arel \\
Department of Electrical Engineering and Computer Science \\
University of Tennessee \\
\texttt{itamar@ieee.org} \\
}

\nipsfinalcopy 

\usepackage{babel}

\makeatother

\begin{document}
\maketitle 
\begin{abstract}
Scalability properties of deep neural networks raise key research
questions, particularly as the problems considered become larger and
more challenging. This paper expands on the idea of conditional computation
introduced in \cite{Deep_Learning_Looking_Forward_2013}, where the
nodes of a deep network are augmented by a set of gating units that
determine when a node should be calculated. By factorizing the weight
matrix into a low-rank approximation, an estimation of the sign of
the pre-nonlinearity activation can be efficiently obtained. For networks
using rectified-linear hidden units, this implies that the computation
of a hidden unit with an estimated negative pre-nonlinearity can be
omitted altogether, as its value will become zero when nonlinearity
is applied. For sparse neural networks, this can result in considerable
speed gains. Experimental results using the MNIST and SVHN data sets
with a fully-connected deep neural network demonstrate the performance
robustness of the proposed scheme with respect to the error introduced
by the conditional computation process. 
\end{abstract}

\section{Introduction}

In recent years, deep neural networks have redefined state-of-the-art
in many application domains, notably in computer vision \cite{Krizhevsky_Imagenet_2012}
and speech processing \cite{Mohamed_TIMIT_2011}. In order to scale
to more challenging problems, however, neural networks must become
larger, which implies an increase in computational resources. Shifting
computation to highly parallel platforms such as GPUs has enabled
the training of massive neural networks that would otherwise train
too slowly on conventional CPUs. While the extremely high computational
power used for the experiment performed in \cite{Large_Scale_Unsupervised_Learning_2012}
(16,000 cores training for many days) was greatly reduced in \cite{Deep_Learning_COTS_HPC_2013}
(3 servers training for many days), specialized high-performance platforms
still require several machines and several days of processing time.
However, there may exist more fundamental changes to the algorithms
involved which can greatly assist in scaling neural networks.

Many of these state-of-the-art networks have several common properties:
the use of rectified-linear activation functions in the hidden neurons,
and a high level of sparsity induced by dropout regularization or
a sparsity-inducing penalty term on the loss function. Given that
many of the activations are effectively zero, due to the combination
of sparsity and the hard thresholding of rectified linear units, a
large amount of computation is wasted on calculating values that are
eventually truncated to zero and provide no contribution to the network
outputs or error components. Here we focus on this key observation
in devising a scheme that can predict the zero-valued activations
in a computationally cost-efficient manner.

\section{Conditional Computation in Deep Neural Networks}

\subsection{Exploiting Redundancy in Deep Architectures}

\begin{figure}
\begin{centering}
\includegraphics[width=4in]{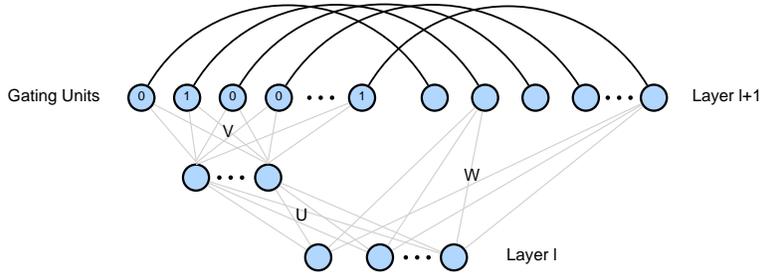} 
\par\end{centering}

\caption{\label{gating_units}An illustration of an activation estimator. $U$
and $V$ represent the factorization of the low-rank matrix and $W$
denotes the full-rank matrix. In this case, the activation estimator
recommends that only the $2^{nd}$ and the $n^{th}$ neuron be computed
for layer $l+1$.}
\end{figure}

In \cite{Denil_Predicting_2013}, the authors made the observation
that deep models tend to have a high degree of redundancy in their
weight parameterization. The authors exploit this redundancy in order
to train as few as 5\% of the weights in a neural network while estimating
the other 95\% with the use of carefully constructed low-rank decompositions
of the weight matrices. Such a reduction in the number of active training
parameters can render optimization easier by reducing the number of
variables to optimize over. Moreover, it can help address the problem
of scalability by greatly reducing the communication overhead in a
distributed system.

Assuming there is a considerable amount of redundancy in the weight
parameterization, a similar level of redundancy is likely found in
the activation patterns of individual neurons. Therefore, given an
input sample, the set of redundant activations in the network may
be approximated. If a sufficiently accurate approximation can be obtained
using low computational resources, activations for a subset of neurons
in the network's hidden layers need not be calculated.

In \cite{Deep_Learning_Looking_Forward_2013} and \cite{bengio_condcomp},
the authors propose the idea of conditional computation in neural
networks, where the network is augmented by a gating model that turns
activations on or off depending on the state of the network. If this
gating model is able to reliably estimate which neurons need to be
calculated for a particular input, great improvements in computational
efficiency may be obtainable if the network is sufficiently sparse.
Figure \ref{gating_units} illustrates a conditional computation unit
augmenting a layer of a neural net by using some function $f\left(U,V,a_{l}\right)$
to determine which hidden unit activations $a_{l+1}$ should be computed
given the activations $a_{l}$ of layer $l$.

\subsection{Sparse Representations, Activation Functions, and Prediction}

In some situations, sparse representations may be superior to dense
representations, particularly in the context of deep architectures
\cite{Deep_Sparse_Rectifier_Networks_2011}. However, sparse representations
learned by neural networks with sigmoidal activations are not truly
``sparse'', as activations only approach zero in the limit towards
negative infinity. A conditional computation model estimating the
sparsity of a sigmoidal network would thus have to impose some threshold,
beyond which the neuron is considered inactive. So-called ``hard-threshold''
activation functions such as rectified-linear units, on the other
hand, produce true zeros which can be used by conditional computation
models without imposing additional hyperparameters.

\section{Problem Formulation}

\subsection{Estimating Activation Sign via Low-Rank Approximation}

\begin{figure}
\begin{centering}
\includegraphics[width=4.5in]{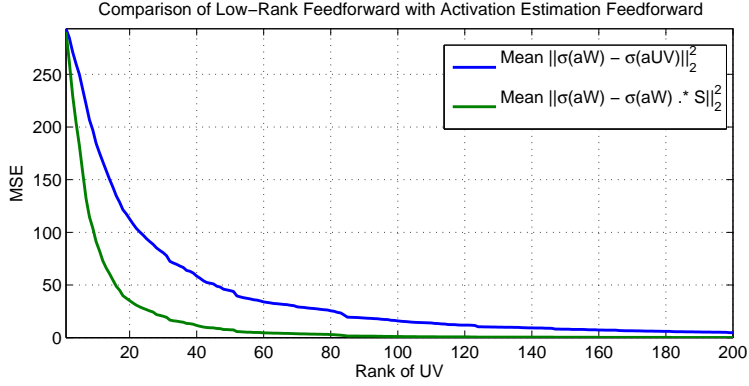} 
\par\end{centering}

\caption{\label{fro-vs-est}The low-rank approximation $UV$ can be substituted
in for $W$, and can approximate the matrix $W$ with a relatively
low rank. However, if we use the output of the activation estimator
$S$, as defined in Eq. (\ref{eq:act_est}), with the full-rank feedforward,
$\sigma\left(aW\right)\cdot S$, a lower-rank approximation can be
utilized. The activations and weights are from the first layer of
a neural network trained on MNIST, and the factorization $UV$ is
obtained via SVD.}
\end{figure}

Given the activation $a_{l}$ of layer $l$ of a neural network, the
activation $a_{l+1}$ of layer $l+1$ is given by:

\begin{equation}
a_{l+1}=\sigma(a_{l}W_{l})\label{l_feedforward}
\end{equation}

where $\sigma(\cdot)$ denotes the function defining the neuron's
non-linearity, $a_{l}\in\mathbb{R}^{n\times h_{1}}$, $a_{l+1}\in\mathbb{R}^{n\times h_{2}}$,
$W_{l}\in\mathbb{R}^{h_{1}\times h_{2}}$. If the weight matrix is
highly redundant, as in \cite{Denil_Predicting_2013}, it can be well-approximated
using a low-rank representation and we may rewrite (\ref{l_feedforward})
as

\begin{equation}
a_{l+1}=\sigma(a_{l}U_{l}V_{l})\label{l_feedforward_lowrank}
\end{equation}

where $U_{l}V_{l}$ is the low-rank approximation of $W_{l}$, $U_{l}\in\mathbb{R}^{h_{1}\times k}$,
$V_{l}\in\mathbb{R}^{k\times h_{2}}$, $k\ll\min(h_{1},h_{2})$. So
long as $k<\frac{h_{1}h_{2}}{h_{1}+h_{2}}$, the low-rank multiplication
$a_{l}U_{l}V_{l}$ requires fewer arithmetic operations than the full-rank
multiplication $a_{l}W_{l}$, assuming the multiplication $a_{l}U_{l}$
occurs first. When $\sigma(\cdot)$ is the rectified-linear function,

\begin{equation}
\sigma(x)=\max(0,x)\label{relu}
\end{equation}

such that all negative elements of the linear transform $a_{l}W_{l}$
become zero, one only needs to estimate the sign of the elements of
the linear transform in order to predict the zero-valued elements.
Assuming the weights in a deep neural network can be well-approximated
using a low-rank estimation, the small error in the low-rank estimation
is of marginal relevance in the context of recovering the sign of
the operation. 

Given a low-rank approximation $W_{l}\approx U_{l}V_{l}=\hat{W}_{l}$,
the estimated sign of $a_{l+1}$ is given by 
\begin{equation}
sgn(a_{l+1})\approx sgn(a_{l}\hat{W}_{l})
\end{equation}
Each element $\left(a_{l+1}\right)_{i,j}$ is given by a dot product
between the row vector $a_{l}^{\left(i\right)}$ and the column vector
$W_{l}^{(j)}$. If $sgn(a_{l}\hat{W}_{l}^{(j)})=-1$, then the true
activation $\left(a_{l+1}\right)_{i,j}$ is likely negative, and will
likely become zero after the rectified-linear function is applied.
Considerable speed gains are possible if we skip those dot products
based on the prediction; such gains are especially substantial when
the network is very sparse. The overall activation for a hidden layer
$l$ augmented by the activation estimator is given by $\sigma\left(a_{l}W_{l}\right)\cdot S_{l}$,
where $\cdot$ denotes the element-wise product and $S_{l}$ denotes
a matrix of zeros and ones, where 
\begin{equation}
\left(S_{l}\right)_{i,j=}\begin{cases}
0, & sgn\left(\left(a_{l}U_{l}V_{l}\right)_{i,j}\right)=-1\\
1, & sgn\left(\left(a_{l}U_{l}V_{l}\right)_{i,j}\right)=+1
\end{cases}\label{eq:act_est}
\end{equation}

Figure \ref{fro-vs-est} illustrates the error profile of a neural
network using the low-rank estimation $UV$ in place of $W$ compared
with a neural network augmented with an activation sign estimator
as the rank is varied from one to full-rank. One can see that the
error of the activation sign estimator diminishes far more quickly
than the error of the low-rank activation, implying that the sign
estimator can do well with a relatively low-rank approximation of
$W$.

\subsection{SVD as a Low-Rank Approximation}

The Singular Value Decomposition (SVD) is a common matrix decomposition
technique that factorizes a matrix $A\in\mathbb{R}^{m\times n}$ into
$A=U\Sigma V^{T}$, $U\in\mathbb{R}^{m\times m},\Sigma\in\mathbb{R}^{m\times n},V\in\mathbb{R}^{n\times n}$.
By \cite{svd_lowrank}, the matrix $A$ can be approximated using
a low rank matrix $\hat{A}_{r}$ corresponding to the solution of
the constrained optimization of 
\begin{equation}
\min_{\hat{A}_{r}}\|A-\hat{A}_{r}\|_{F}
\end{equation}
where $\|\cdot\|_{F}$ is the Frobenius norm, and $\hat{A}_{r}$ is
constrained to be of rank $r<rank(A)$. The minimizer $\hat{A}_{r}$
is given by taking the first $r$ columns of $U$, the first $r$
diagonal entries of $\Sigma$, and the first $r$ columns of $V$.
The resulting matrices $U_{r}$, $\Sigma_{r}$, and $V_{r}$ are multiplied,
yielding $\hat{A}_{r}=U_{r}\Sigma_{r}V_{r}^{T}$. The low-rank approximation
$\hat{W}=UV$ is then defined such that $\hat{W}=U_{r}(\Sigma_{r}V_{r}^{T})$,
where $U=U_{r}$ and $V=\Sigma_{r}V_{r}^{T}$.

Unfortunately, calculating the SVD is an expensive operation, on the
order of $O(mn^{2})$, so recalculating the SVD upon the completion
of every minibatch adds significant overhead to the training procedure.
Given that we are uniquely interested in estimating in the sign of
$a_{l+1}=a_{l}W_{l}$, we can opt to calculate the SVD less frequently
than once per minibatch, assuming that the weights $W_{l}$ do not
change significantly over the course of a single epoch so as to corrupt
the sign estimation.

\subsection{Encouraging Neural Network Sparsity}

To overcome the additional overhead imposed by the conditional computation
architecture, the neural network must have sparse activations. Without
encouragement to settle on weights that result in sparse activations,
such as penalties on the loss function, a neural network will not
necessarily become sparse enough to be useful in the context of conditional
computation. Therefore, an $\ell_{1}$ penalty for the activation
vector of each layer is applied to the overall loss function, such
that 
\begin{equation}
J(W,\lambda)=L(W)+\lambda\sum_{l=1}^{L}\|a_{l}\|_{1}
\end{equation}
Such a penalty is commonly used in sparse dictionary learning algorithms
and tends to push elements of $a_{l}$ towards zero \cite{lee2006efficient}.

Dropout regularization \cite{Dropout_Co_Adaptation_2012} is another
technique known to sparsify the hidden activations in a neural network.
Dropout first sets the hidden activations $a_{l}$ to zero with probability
$p$. During training, the number of active neurons is likely less
than $p$ for each minibatch. When the regularized network is running
in the inference mode, dropout has been observed to have a sparsifying
effect on the hidden activations \cite{srivastava_dropout_ms}. The
adaptive dropout method \cite{adaptive_dropout} can further decrease
the number of active neurons without degrading the performance of
the network.

\subsection{Theoretical Speed Gain}

For every input example, a standard neural network computes $\sigma\left(aW\right)$,
where $a\in\mathbb{R}^{N\times d}$ and $W\in\mathbb{R}^{d\times h}$,
where $N=1$ for a fully-connected network, or $N$ is the number
of convolutions for a convolutional network. Assuming additions and
multiplications are constant-time operations, the matrix multiplication
requires $N\left(2d-1\right)h$ floating point operations (we need
to compute $Nh$ dot products, where each dot product consists of
$d$ multiplications and $d-1$ additions), and the activation function
requires $Nh$ floating point operations, yielding $N\left(2d-1\right)h+Nh$
operations. The activation estimator $\sigma\left(aUV\right)$, $U\in\mathbb{R}^{d\times k}$,
$V\in\mathbb{R}^{k\times h}$ requires $N\left(2d-1\right)k+N\left(2k-1\right)h$
floating point operations for the low-rank multiplication followed
by $Nh$ operations for the $sgn\left(\cdot\right)$ activation function,
yielding $N\left(2d-1\right)k+N\left(2k-1\right)h+Nh$. However, given
a sparsity coefficient $\alpha\in\left[0,1\right]$ (where $\alpha=0$
implies no activations are active, and $\alpha=1$ implies all activations
are active), a conditional matrix multiplication would require $\alpha N\left(2d-1\right)h+\alpha Nh$
operations. The SVD calculation to obtain the activation estimation
weights is $\beta O\left(nd\min\left(n,d\right)\right)$, where $\beta$
is the ratio of feed-forwards to SVD updates (eg., with a minibatch
size of 250, a training set size of 50,000, and once-per-epoch SVD
updates, $\beta=\frac{250}{50000}=0.005$).

Altogether, the number of floating point operations for calculating
the feed-forward in a layer in a standard neural network is
\begin{equation}
F_{nn}=N\left(2d-1\right)h+Nh
\end{equation}
and the number of floating point operations for the activation estimation
network with conditional computation is
\begin{equation}
F_{ae}=N\left(2d-1\right)k+N\left(2k-1\right)h+Nh+\alpha h\left(N\left(2d-1\right)h+Nh\right)+\beta O\left(nd\min\left(n,d\right)\right)
\end{equation}
The relative reduction of floating point operations for a layer can
be represented as $\frac{F_{nn}}{F_{ae}}$, and is simplified as
\begin{equation}
\frac{2dh}{k\left(2d+2h-1\right)+2\alpha dh+\beta O\left(nd\min\left(n,d\right)\right)}\label{eq:reduction}
\end{equation}

For a neural network with many layers, the relative speedup is given
by
\begin{equation}
\frac{{\displaystyle \sum_{i=1}^{L}F_{nn}^{\left(l\right)}}}{{\displaystyle \sum_{i=1}^{L}F_{ae}^{\left(l\right)}}}
\end{equation}
where ${\displaystyle F_{nn}^{\left(l\right)}}$ is the number of
floating point operations for the $l^{th}$ layer of the full network,
and ${\displaystyle F_{ae}^{\left(l\right)}}$ is the number of floating
point operations for the $l^{th}$ layer of the network augmented
by the activation estimation network. The overall speedup is greatly
dependent on the sparsity of the network and the overhead of the activation
estimator.

\subsection{Implementation Details}

The neural network is built using Rasmus Berg Palm's Deep Learning
Toolbox \cite{RB_Palm_DeepLearnToolbox}. All hidden units are rectified-linear,
and the output units are softmax trained with a negative log-likelihood
loss function. The weights, $w$, are initialized as $w\sim\mathcal{N}\left(0,\sigma^{2}\right)$
and biases $b$ are set to 1 in order to encourage the neurons to
operate in their non-saturated region once training begins, as suggested
in \cite{Krizhevsky_Imagenet_2012}. In all experiments, the dropout
probability $p$ is fixed to 0.5 for the hidden layers.

The learning rate $\gamma$ is scheduled such that $\gamma_{n}=\gamma_{0}\lambda^{n}$
where $\gamma_{n}$ is the learning rate for the $n^{th}$ epoch,
$\gamma_{0}$ is the initial learning rate, and $\lambda$ is a decay
term slightly less than 1, eg., 0.995. The momentum term $\nu$ is
scheduled such that $\nu_{n}=\max\left(\nu_{max},\nu_{0}\beta^{n}\right)$
where $\nu_{n}$ is the momentum for the $n^{th}$ epoch, $\nu_{max}$
is the maximum allowed momentum, $\nu_{0}$ is the initial momentum,
and $\beta$ is an incremental term slightly greater than 1, eg.,
1.05.

To simplify prototyping, the feed-forward is calculated for a layer,
and the activation estimator is immediately applied before the next
layer activations are used. This is equivalent to bypassing the calculations
for activations that are likely to produce zeros. In practice, re-calculating
the SVD once per epoch for the activation estimator seems to be a
decent tradeoff between activation estimation accuracy and computational
efficiency, but this may not necessarily be true for other datasets.

\section{Experimental Results}

\subsection{SVHN}

\begin{table}
\begin{centering}
\begin{tabular}{|c|c|c|}
\cline{2-3} 
\multicolumn{1}{c|}{} & \textbf{SVHN}  & \textbf{MNIST}\tabularnewline
\hline 
\emph{Architecture}  & 1024-1500-700-400-200-10  & 784-1000-600-400-10\tabularnewline
\hline 
\emph{Weight Init}  & $w\sim\mathcal{N}\left(0,0.01\right)$; $b=1$  & $w\sim\mathcal{N}\left(0,0.05\right)$; $b=1$\tabularnewline
\hline 
\emph{Init Learning Rate}  & 0.15  & 0.25\tabularnewline
\hline 
\emph{Learning Rate Scaling}  & 0.99  & 0.99\tabularnewline
\hline 
\emph{Maximum Momentum}  & 0.8  & 0.8\tabularnewline
\hline 
\emph{Momentum Increment}  & 1.01  & 1.05\tabularnewline
\hline 
\emph{Maximum Norm}  & 25  & 25\tabularnewline
\hline 
\emph{$\ell_{1}$ Activation Penalty}  & 0  & $1\times10^{-5}$\tabularnewline
\hline 
\emph{$\ell_{2}$Weight Penalty}  & -- & $5\times10^{-5}$\tabularnewline
\hline 
\end{tabular}
\par\end{centering}

\caption{\label{hyperparams}Hyperparameters for SVHN and MNIST experiments.}
\end{table}

\begin{figure}
\begin{centering}
\includegraphics[width=4in]{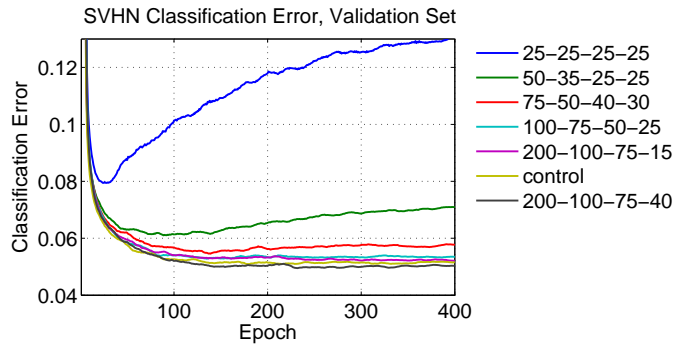} 
\par\end{centering}

\caption{\label{svhn-runs}Classification error of the validation set for SVHN
on seven configurations of the activation estimator for each hidden
layer. The 'control' network has no activation estimator and is used
as a baseline of comparison for the other networks.}
\end{figure}

Street View House Numbers (SVHN) \cite{netzer2011reading} is a large
image dataset containing over 600,000 labeled examples of digits taken
from street signs. Each example is an RGB $32\times32$ (3072-dimensional)
image. To pre-process the dataset, each image is transformed into
the YUV colorspace. Next, local contrast normalization \cite{jarrett2009best}
followed by a histogram equalization is applied to the Y channel.
The U and V channels are discarded, resulting in a 1024-dimensional
vector per example. The dataset is then normalized for the neural
network by subtracting out the mean and dividing by the square root
of the variance for each variable. To select the hyperparameters,
the training data was split into 590,000 samples for the training
set and 14,388 samples for the validation set. The architecture was
held fixed while the other hyperparameters were chosen randomly over
30 runs using a network with no activation estimation. The hyperparameters
of the neural network with the lowest resulting validation error were
then used for all experiments.

To evaluate the sensitivity of the activation estimator, several parameterizations
for the activation estimator are evaluated. Each network is trained
with the hyperparameters in Table \ref{hyperparams}, and the results
of seven parameterizations are shown in Figure \ref{svhn-runs}. Each
parameterization is described by the rank of each approximation, eg.,
`75-50-40-30' describes a network with an activation estimator using
a 75-rank approximation for $W_{1}$, a 50-rank approximation for
$W_{2}$, a 40-rank approximation for $W_{3}$, and a 30-rank approximation
for $W_{4}$. Note that a low-rank approximation is not necessary
for $W_{5}$ (the weights connecting the last hidden layer to the
output layer), as we do not want to approximate the activations for
the output layer.

\begin{figure}
\begin{centering}
\includegraphics[width=4in]{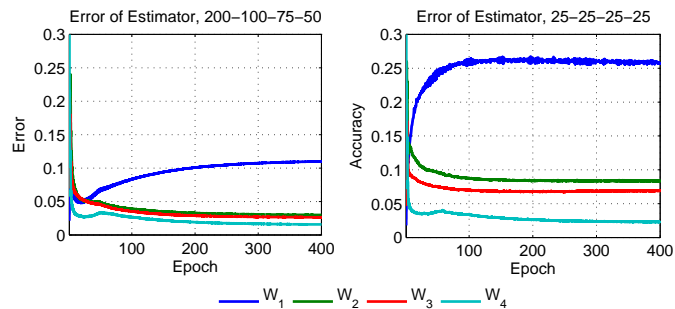} 
\par\end{centering}

\caption{\label{svhn-accest}A comparison of a low-rank activation estimator
and a higher-rank activation estimator. In this instance, a 25-25-25-25
activation estimator is too coarse to adequately capture the structure
of the weight matrices.}
\end{figure}

Some runs, specifically 25-25-25-25 and 50-35-25-25 in Figure \ref{svhn-runs}
exhibit an initial decrease in classification error, followed by a
gradual increase in classification error as training progresses. In
the initial epochs, the hidden layer activations are mostly positive
because the weights are relatively small and the biases are very large.
As a consequence, the activation estimation is a much simpler task
for the initial epochs. However, as the pattern of the activation
signs diversifies as the network continues to train, the lower-rank
approximations begin to fail, as illustrated in Figure \ref{svhn-accest}.

Table \ref{svhn-test} summarizes the test set error for the control
and activation estimation networks. $W_{1}$ appears to be most sensitive,
quickly reducing the test set error from 10.72\% to 12.16\% when the
rank of $\hat{W}_{1}$ is lowered from 75 to 50. The rank of $\hat{W}_{4}$
appears to be the least sensitive, reducing the test set error from
9.96\% to 10.01\% as the rank is lowered from 25 to 15.

\begin{table}
\begin{centering}
\begin{tabular}{|c|c|}
\hline 
\textbf{Network}  & \textbf{Error}\tabularnewline
\hline 
Control  & 9.31\%\tabularnewline
\hline 
200-100-75-15  & 9.67\%\tabularnewline
\hline 
100-75-50-25  & 9.96\%\tabularnewline
\hline 
100-75-50-15  & 10.01\%\tabularnewline
\hline 
75-50-40-30  & 10.72\%\tabularnewline
\hline 
50-40-40-35  & 12.16\%\tabularnewline
\hline 
25-25-15-15  & 19.40\%\tabularnewline
\hline 
\end{tabular}
\par\end{centering}

\caption{\label{svhn-test}SVHN test set error for seven networks.}
\end{table}

\subsection{MNIST}

\begin{figure}
\begin{centering}
\includegraphics[width=4in]{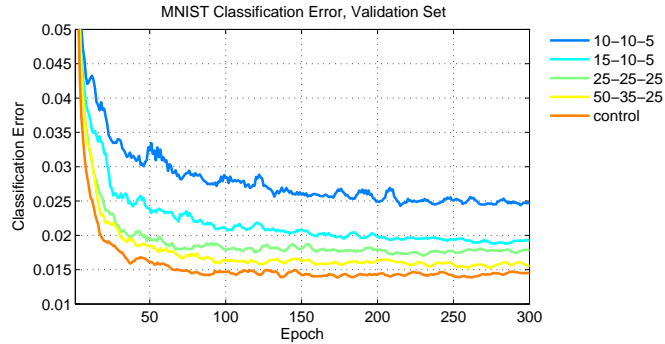} 
\par\end{centering}

\caption{\label{mnist-runs}Classification error of the validation set for
MNIST on five configurations of the activation estimator for each
hidden layer.}
\end{figure}

MNIST is a well-known dataset of hand-written digits containing 70,000
$28\times28$ labeled images, and is generally split into 60,000 training
and 10,000 testing examples. Very little pre-processing is required
to achieve good results - each feature is transformed by $x_{t}=\frac{x}{\sqrt{\sigma_{max}^{2}}}-0.5$,
where $x$ is the input feature, $\sigma_{max}^{2}$ is the maximum
variance of all features, and 0.5 is a constant term to roughly center
each feature. To select the hyperparameters, the training data was
split into 50,000 samples for the training set and 10,000 samples
for the validation set. The architecture was held fixed while the
other hyperparameters were chosen randomly over 30 runs using a network
with no activation estimation. The hyperparameters of the neural network
with the lowest resulting validation error were then used for all
experiments. Several parameterizations for the activation estimator
are evaluated for a neural network trained with the hyperparameters
listed in Table \ref{hyperparams} using the same approach as the
SVHN experiment above. The results for the validation set plotted
against the epoch number are shown in Figure \ref{mnist-runs}, and
the final test set accuracy is reported in Table \ref{mnist-test}.

A neural network with a very low-rank weight matrix in the activation
estimation can train surprisingly well on MNIST. Lowering the rank
from 784-600-400 to 50-35-25 impacts performance negligibly. Ranks
as low as 25-25-25 does not lessen performance too greatly, and ranks
as low as 10-10-5 yield a classifier capable of 2.28\% error.

\begin{table}
\begin{centering}
\begin{tabular}{|c|c|}
\hline 
\textbf{Network}  & \textbf{Error}\tabularnewline
\hline 
Control  & 1.40\%\tabularnewline
\hline 
50-35-25  & 1.43\%\tabularnewline
\hline 
25-25-25  & 1.60\%\tabularnewline
\hline 
15-10-5  & 1.85\%\tabularnewline
\hline 
10-10-5  & 2.28\%\tabularnewline
\hline 
\end{tabular}
\par\end{centering}

\caption{\label{mnist-test}MNIST test set error for five networks.}
\end{table}

\section{Discussion and Further Work}

Low-rank estimations of weight matrices of a neural network obtained
via once-per-epoch SVD work very well as efficient estimators of the
sign of the activation for the next hidden layer. In the context of
rectified-linear hidden units, computation time can be reduced greatly
if this estimation is reliable and the hidden activations are sufficiently
sparse. This approach is applicable to any hard-thresholding activation
function, such as the functions investigated in \cite{goroshin2013saturating},
and can be easily extended to be used with convolutional neural networks.

\begin{figure}
\begin{centering}
\includegraphics[width=4in]{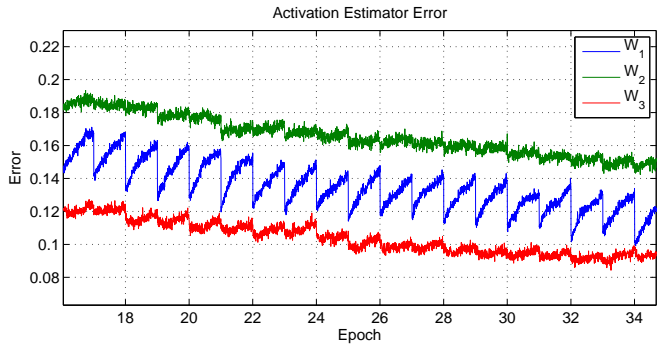} 
\par\end{centering}

\caption{\label{svd_zigs}Because the SVD is calculated at the beginning of
each epoch, each subsequent gradient update in each minibatch moves
the weight matrix further from low-rank factorization, resulting in
an increasing error until the SVD is recalculated at the beginning
of the next epoch. Different layers are negatively impacted in differing
degrees.}
\end{figure}

While the activation estimation error does not tend to deviate too
greatly inbetween minibatches over an epoch, as illustrated in Figure
\ref{svd_zigs}, this is not guaranteed. An online approach to the
low-rank approximation would therefore be preferable to a once-per-epoch
calculation. In addition, while the low-rank approximation given by
SVD minimizes the objective function $\|A-\hat{A}_{r}\|_{F}$, this
is not necessarily the best objective function for an activation estimator,
where we seek to minimize $\left\Vert \sigma\left(aW\right)-\sigma\left(aW\cdot S\right)\right\Vert $,
which is a much more difficult and non-convex objective function.
Also, setting the hyperparameters for the activation estimator can
be a tedious process involving expensive cross-validation when an
adaptive algorithm could instead choose the rank based on the spectrum
of the singular values. Therefore, developing a more suitable low-rank
approximation algorithm could provide a promising future direction
of research.

In \cite{adaptive_dropout}, the authors propose a method called ``adaptive
dropout'' by which the dropout probabilities are chosen by a function
optimized by gradient descent instead of fixed to some value. This
approach bears some resemblance to this paper, but with the key difference
that the approach in \cite{adaptive_dropout} is motivated by improved
regularization and this paper's method is motivated by computational
efficiency. However, the authors introduce a biasing term that allows
for greater sparsity that could be introduced into this paper's methodology.
By modifying the conditional computation unit to compute $sgn\left(aUV-b\right)$,
where $b$ is some bias, we can introduce a parameter that can tune
the sparsity of the network, allowing for a more powerful trade-off
between accuracy and computational efficiency.

\subsection*{Acknowledgments}

This work was partially supported by the Defense Advanced Research
Projects Agency (DARPA) under contract number HR0011-13-2-0016.

 \bibliographystyle{plain}
\bibliography{master}

\end{document}